\documentclass[runningheads]{llncs}
\usepackage[T1]{fontenc}
\usepackage{graphicx,verbatim}
\usepackage[table,xcdraw]{xcolor}
\usepackage{xcolor}
\usepackage[dvipsnames]{xcolor}
\definecolor{mygreen}{RGB}{114, 240, 126}
\usepackage{multirow}
\usepackage{amsmath}
\usepackage{amssymb}
\usepackage{booktabs}
\usepackage{pifont}
\usepackage{enumitem}
\usepackage{colortbl}
\usepackage{subcaption}
\usepackage{bbding}
\usepackage{hyperref} 
\usepackage[capitalize]{cleveref}
\usepackage{cite}
\crefname{section}{Sec.}{Secs.}
\Crefname{section}{Section}{Sections}
\Crefname{table}{Table}{Tables}
\crefname{table}{Tab.}{Tabs.}
\makeatletter
\def\@fnsymbol#1{
  \ifcase#1\or
    *\or
    \ensuremath{\dagger}\or
    \ensuremath{\ddagger}\or
    \ensuremath{\mathsection}\or
    \ensuremath{\paragraph}\or
    \ensuremath{\|}\or
    **\or
    \ensuremath{\dagger\dagger}\or
    \ensuremath{\ddagger\ddagger}
  \else
    \@ctrerr
  \fi
}

\begin{document}

\title{VA-Adapter: Adapting Ultrasound Foundation Model to Echocardiography Probe Guidance}
\titlerunning{VA-Adapter for Echocardiography Probe Guidance}

\author{
Teng Wang\inst{1}\textsuperscript{*}
\and
Haojun Jiang\inst{1}\textsuperscript{*}\textsuperscript{\ensuremath{\ddag}}
\and
Yuxuan Wang\inst{2}
\and
Zhenguo Sun\inst{3}
\and
Yujiao Deng\inst{4}
\and
Shiji Song\inst{1}
\and
Gao Huang\inst{1}\textsuperscript{\ensuremath{\dagger}}
}

\authorrunning{Wang et al.}

\institute{
Department of Automation, BNRist, Tsinghua University, Beijing, China
\and
School of Computer Science and Technology, Xidian University,
Xi’an, China
\and
Beijing Academy of Artificial Intelligence, Beijing, China
\and
Chinese PLA General Hospital, Beijing, China 
\\
\email{\{t-wang25, jhj20\}@mails.tsinghua.edu.cn, gaohuang@tsinghua.edu.cn}
}

\maketitle              

\begingroup
\renewcommand{\thefootnote}{\fnsymbol{footnote}}
\setcounter{footnote}{0}
\footnotetext[1]{Equal contribution. \quad \(\ddagger\) Guided this work. \quad \(\dagger\) Corresponding author.}
\endgroup

\begin{abstract}

Echocardiography is a critical tool for detecting heart diseases, yet its steep operational difficulty causes a shortage of skilled personnel. 
Probe guidance systems, which assist in acquiring high-quality images, offer a promising solution to lower this operational barrier.
However, robust probe guidance remains challenging due to significant individual variability. 
This variability manifests as differences in low-level features within two-dimensional (2D) images, which complicates image feature understanding, and differences in individual three-dimensional (3D) structures, which poses challenges for precise navigation.
To address these challenges, we first propose leveraging the robust image representations learned by ultrasound foundation models from vast datasets.
Yet, applying these models to probe navigation is non-trivial due to their lack of understanding of individual 3D structures.
To this end, we meticulously design a Vision-Action Adapter (VA-Adapter) to online inject the capability of understanding individual 3D structures.
Specifically, by embedding the VA-Adapter into the foundation model's image encoder, the model can infer cardiac anatomy from historical vision-action sequences, mimicking the cognitive process of a sonographer.
Extensive experiments on a dataset with over 1.31M samples demonstrate that the VA-Adapter outperforms strong probe guidance models while requiring approximately 33 times fewer trained parameters.
Code is available at https://github.com/LeapLabTHU/VA-Adapter.


\end{abstract}

\section{Introduction}
\label{sec:introduction}

\begin{figure*}[t]
\centering
\includegraphics[width=\textwidth]{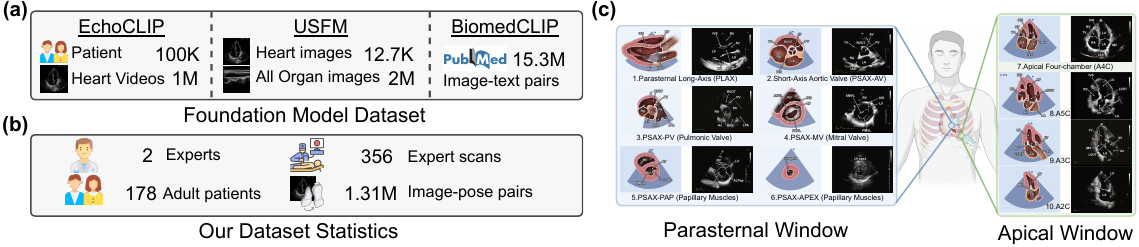}
\caption{
Illustration of the dataset.
(a) Large-scale diagnostic foundation model dataset.
(b) Our dataset statistic.
(c) Standard planes (images from \cite{mitchell2019guidelines}).
}
\label{fig:dataset}
\vspace{-2pt}
\end{figure*}

Cardiovascular diseases have become a significant factor affecting human lifespan~\cite{roth2017global}.  
Echocardiography is a commonly used imaging technique for diagnosing cardiovascular diseases, allowing the observation of the health conditions of heart chambers, valves, and blood vessels~\cite{mitchell2019guidelines}.  
With technological advancements, AI-driven echocardiography diagnostic models~\cite{christensen2024vision,ghorbani2020deep,ouyang2020video} have demonstrated remarkable capabilities.  
As shown in \cref{fig:dataset}(a), representative models include EchoCLIP~\cite{christensen2024vision}, pre-trained on over one million paired cardiac ultrasound videos and expert reports; USFM~\cite{jiao2024usfm}, pre-trained on over two million ultrasound images across 12 organs; and BiomedCLIP~\cite{zhang2023biomedclip}, pre-trained on 15.3 million image–text pairs from 4.4 million PubMed articles. All three models demonstrate strong capabilities in interpreting cardiac ultrasound images.

Undoubtedly, the prerequisite for these powerful diagnostic models to function effectively is the availability of high-quality ultrasound images.  
However, due to the inherently high operational difficulty of cardiac ultrasound, it takes years of training for a beginner to master the technique, resulting in a scarcity of skilled professionals.
Therefore, leveraging AI technology to assist in cardiac ultrasound scanning is a crucial research direction.

In recent years, researchers~\cite{bao2024real,droste2020automatic,jiang2024structure,jiang2024cardiac,jiang2025ultrasep,jiang2025towards,li2023rl,narang2021utility,wang2025ultrahit,yue2025echoworld} have developed AI-driven probe guidance systems to acquire higher-quality ultrasound images. Droste et al.~\cite{droste2020automatic} proposed US-GuideNet for fetal plane scanning. Narang et al.~\cite{narang2021utility} collected large-scale echocardiography scanning data and trained a CNN-based guidance system from scratch, but it is commercial and closed-source. Li et al.~\cite{li2023rl} used cardiac CT as a simulation environment and applied reinforcement learning to learn guidance policies.
Despite progress, these works study probe guidance separately from diagnosis and do not leverage advances in diagnostic foundation models~\cite{christensen2024vision,jiao2024usfm,zhang2023biomedclip} (e.g., EchoCLIP). Since both scanning and diagnosis require understanding cardiac ultrasound structures and making decisions, we hypothesize that diagnostic-model advances can also improve probe guidance.

In this paper, we aim to build upon the ultrasound foundation model's ability to interpret cardiac ultrasound images by equipping it with the capability to understand three-dimensional (3D) cardiac structures and reason about probe adjustment actions.
First, to preserve the basic capabilities learned by foundation model from large-scale data as much as possible, we employ a parameter-efficient fine-tuning strategy called Adapter, which freezes the foundation model's image encoder and only optimizes the parameters within the adapter.
Further, to better exploit individual 3D anatomy, we propose a Vision–Action Adapter (VA-Adapter) that encodes vision–action sequences and learns individual-specific cardiac structure from them.
Specifically, VA-Adapters are inserted into deeper encoder layers of foundation model, where features are more task-relevant~\cite{yang2024mma}.
Finally, extensive experiments show that the VA-Adapter equips diagnostic foundation models with better probe guidance capability at a low training cost.

\section{Method}

\begin{figure*}[t]
\centering
\includegraphics[width=\textwidth]{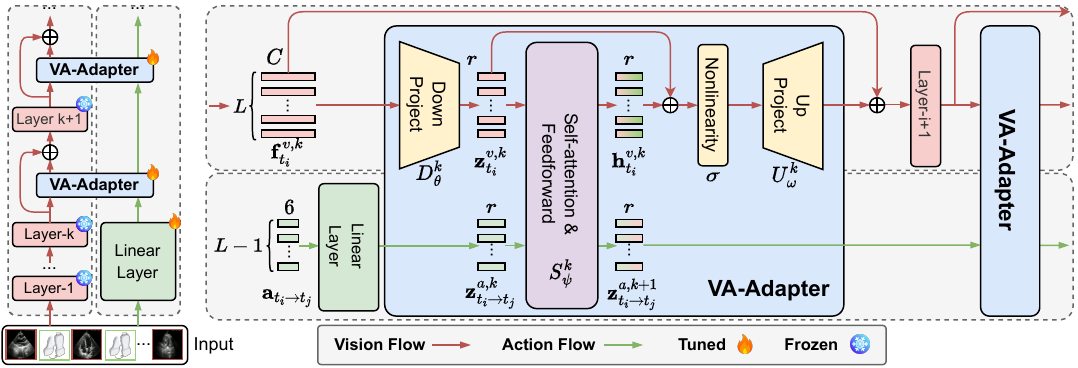}
\caption{
Illustration of the architecture of the VA-Adapter.
The left side shows that we insert VA-Adapter into the deep layers of foundation models, and the right side shows the internal structure of VA-Adapter.
}
\label{fig:adapter}
\end{figure*}

In this section, We first describe the cardiac ultrasound scanning dataset, then introduce our vision–action adapter that enables ultrasound foundation models to reason about probe adjustments for target-plane navigation.

\subsection{Dataset Acquisition}
\label{method:dataset}
The dataset used in our work was collected from 178 adult subjects and includes 356 expert scanning trajectories, totaling 1.31 million image-action pairs. 
The data was gathered by two senior sonographers with over 10 years of experience. 
They performed continuous scans of 10 standard echocardiographic planes (\cref{fig:dataset}(c)) using an ultrasound probe attached to the end of a robotic arm. 
The data collection system recorded real-time images acquired by the probe and the corresponding probe pose data. 
This created a scanning sequence $\{(\mathbf{I}_t, \mathbf{p}_t)\}^{T}_{t=1}$, where $\mathbf{I}_t \in\mathbb{R}^{3 \times H \times W} $ is the ultrasound image at time $t$, and $\mathbf{p}_t \in \mathbb{SE}(3)$ is the corresponding probe's 6D pose (3D position $\mathbf{P} \in \mathbb{R}^3$ and 3D orientation $\mathbf{R} \in \mathbb{SO}(3)$).
During the scan, the sonographer marked the standard views. 
Then we can calculate the relative motion of any image $\mathbf{I}_i$ to the standard view $\mathbf{I}_j$ : $\mathbf{a}_{i\to j}=\mathbf{T}_{p_i}^{-1}\mathbf{T}_{p_j}$, where $\mathbf{T}_{p_i}$ and $\mathbf{T}_{p_j}$ represent the transformation matrices corresponding to the probe poses $\mathbf{p}_i$ and $\mathbf{p}_j$. 
Then we use this motion value as the supervisory signal for the ultrasound probe guidance task. 
Notably, sonographers sometimes pause at a probe position to observe cardiac dynamics. As a result, the dataset includes many images from the same probe position but different cardiac phases. These frames share the same motion label, providing implicit supervision that encourages robustness to cardiac-cycle variation.

\subsection{Vision-Action Adapter}
\label{method:adapter}

Echocardiography probe guidance is an emerging area, but progress is limited by data-collection challenges. In contrast, ultrasound image understanding and diagnosis have advanced rapidly, with recent foundation models achieving strong baseline performance. Since probe guidance also requires understanding ultrasound structures, we propose the Vision-Action Adapter to leverage these foundation models for probe guidance. Next, we detail our approach in \cref{fig:adapter}.

\textbf{Input.}
Given a frame $\mathbf{I}_t \in \mathbb{R}^{3 \times H \times W}$ from a scan, we construct an input sequence of length $L$ via segmental sampling~\cite{wang2016temporal}.
Specifically, we split the preceding trajectory into $L{-}1$ equal temporal segments and randomly sample one frame (and its pose) from each segment. The sampled frames are sorted by timestamp and concatenated with the current frame $\mathbf{I}_t$ as the last element $\mathbf{I}_{t_L}$.
Compared with using $L$ consecutive frames, segmental sampling yields larger inter-frame motion and more diverse anatomical viewpoints, which helps the model capture richer 3D structural cues.
For reference, segmental sampling yields an average absolute inter-frame motion of 
$[21.8,\,18.7,\,14.0]\,\mathrm{mm}$ and $[12.6,\,23.3,\,40.9]^\circ$ between adjacent frames, 
indicating substantial variation in both position and orientation.
We then compute relative actions $\mathbf{a}_{t_i \rightarrow t_j}\in\mathbb{R}^{6}$ from the corresponding poses, forming the input trajectory:
    $[\mathbf{I}_{t_1}, \mathbf{a}_{t_1 \rightarrow t_2}, \cdots, \mathbf{I}_{t_{L-1}}, \mathbf{a}_{t_{L-1} \rightarrow t_L}, \mathbf{I}_{t_L}].$

\textbf{Forward Propagation.}
We insert VA-Adapters into the latter part of the foundation model’s vision encoder, where features are more task-relevant~\cite{yang2024mma}. 
For CNN encoders (e.g., EchoCLIP), VA-Adapters are placed between late-stage blocks; for Transformer encoders (e.g., USFM, BiomedCLIP), two VA-Adapters are inserted in each selected block (after attention and after MLP). 
During training, only VA-Adapters are updated while the backbone remains frozen to preserve its learned knowledge.
Given an image $\mathbf{I}_{t_i}$, after the first $k$ vision layers we obtain $\mathbf{f}_{t_i}^{v,k}\in\mathbb{R}^{C}$ (for CNNs, global average pooling is applied). 
We project visual features and actions to the bottleneck space and add timestep embeddings:
\begin{align}
\mathbf{z}_{t_i}^{v,k}=D_\theta^{k}(\mathbf{f}_{t_i}^{v,k})+\mathbf{T}_{i},\quad
\mathbf{z}_{t_i\rightarrow t_j}^{a,k}=A_\phi^{k}(\mathbf{a}_{t_i\rightarrow t_j})+\mathbf{T}_{i},\quad
\mathbf{z}_{t_i}^{v,k},\mathbf{z}_{t_i\rightarrow t_j}^{a,k}\in\mathbb{R}^{r},
\end{align}
where $\mathbf{T}\in\mathbb{R}^{L\times r}$ is timestep embedding, and $r$ is the bottleneck dimension. 
The interleaved vision--action tokens are then processed by a vision--action interaction module $S_\psi^{k}$, designed to extract underlying cardiac structure information:
\begin{align}
[\mathbf{h}_{t_1}^{v,k},\mathbf{z}_{t_1\rightarrow t_2}^{a,k+1},\cdots,\mathbf{h}_{t_L}^{v,k}]
=
S_\psi^{k}([\mathbf{z}_{t_1}^{v,k},\mathbf{z}_{t_1\rightarrow t_2}^{a,k},\cdots,\mathbf{z}_{t_{L-1}}^{v,k},\mathbf{z}_{t_{L-1}\rightarrow t_L}^{a,k},\mathbf{z}_{t_L}^{v,k}]).
\end{align}
The action tokens are forwarded to the next adapter layer, while visual tokens are mapped back to the backbone feature space with a residual connection:
\begin{align}
\mathbf{z}_{t_i}^{v,k+1}=U_\omega^{k}\!\left(\sigma(\mathbf{h}_{t_i}^{v,k}+\mathbf{z}_{t_i}^{v,k})\right)+\mathbf{f}_{t_i}^{v,k}.
\end{align}

\subsection{Task Prediction Head}
\label{method:prediction_head}

After passing through all layers (assuming the network has $M$ layers in total), we obtain the final interleaved sequence features:
\begin{align}
    \mathbf{Z}^{M} = [\mathbf{z}_{t_1}^{v,M}, \mathbf{z}_{t_1 \rightarrow t_2}^{a,M}, \cdots,  
    \mathbf{z}_{t_{L-1}}^{v,M}, \mathbf{z}_{t_{L-1} \rightarrow t_{L}}^{a,M}, \mathbf{z}_{t_L}^{v,M}].
\end{align}
We then apply a GRU-based sequence encoder $E_\rho$ to further aggregate sequential information, and use ten prediction heads $\{H_{\xi_i}\}_{i=1}^{10}$ to predict the action toward each standard plane:
\begin{align}
    \mathbf{V} = E_\rho(\mathbf{Z}^{M}), \ \
    \mathbf{a}^{'}_{t_L \rightarrow t_i} = H_{\xi_i}(\mathbf{v}_{t_L}),
\end{align}
where $\mathbf{V}=[\mathbf{v}_{t_1},\ldots,\mathbf{v}_{t_L}]$ denotes the GRU outputs, and $\mathbf{v}_{t_L}$ corresponds to the current frame $\mathbf{I}_{t_L}$.
Finally, the loss is calculated using the Smooth L1 Loss between the predicted action and the target $\mathbf{a}_{t_L \rightarrow t_{i\text{-th standard plane}}}$ as follows:
\begin{align}
    \mathcal{L}&= \mathcal{L}_{\mathrm{SmoothL1}}(\mathbf{a}_{t_L \rightarrow t_{i\text{-th standard plane}}} , \mathbf{a}^{'}_{t_L \rightarrow t_{i\text{-th standard plane}}} ).
\label{eq:loss}
\end{align}
In \eqref{eq:loss}, translation and rotation are equally weighted; we normalize units in preprocessing (mm for translation, degrees for rotation) to match magnitudes.

\section{Experiments}

\subsection{Datasets and Implementation Details}

\textbf{Datasets.}
Data were collected with a GE machine (\textit{General Electric, USA}) equipped with a M5S probe (\Cref{method:dataset}) under approval and supervision of the University Medical Ethics Committee. We use 284 scans for training and 72 for validation, with data from different subjects in each set.

\textbf{Model Architecture.}
By default, the input sequence length is $L{=}4$ and the adapter bottleneck dimension (adapter dimension) is $r{=}64$, with ReLU activation. The core vision--action interaction module $S_\psi$ is a Transformer block with 4 attention heads and an MLP ratio of 2, using 1D sincos positional encoding. We initialize linear/conv weights with truncated normal ($\sigma{=}0.02$) and set biases to 0; LayerNorm weights are initialized to 1 and biases to 0. 
After the image encoder, image and action features are each projected to 128-d and concatenated as input to the GRU sequence encoder $E_\rho$ (input 256, hidden 128). Each prediction head $H_{\xi_i}$ is a two-layer MLP with GELU in between.

\textbf{Training Strategy.}
We train with Adam (batch size 256) for 5 epochs, using a cosine-decayed learning rate from $1\times10^{-4}$ to $1\times10^{-6}$, and all experiments are performed on four A100 GPUs.

\textbf{Evaluation Metric.}
We report trainable parameters and the Mean Absolute Error (MAE) between predicted $\mathbf{a}'$ and ground-truth $\mathbf{a}$, computed separately for translation and rotation:
\begin{align}
    \text{MAE of Trans.} = \frac{1}{3} \sum_{i=1}^{3} |\mathbf{a}_i - \mathbf{a}^{'}_i|, \ \ \ 
    \text{MAE of Rot.} = \frac{1}{3} \sum_{i=4}^{6} |\mathbf{a}_i - \mathbf{a}^{'}_i|.
\end{align}

\begin{table*}[!t]\normalsize
\caption{
\textbf{Comparison with baseline methods in terms of translation and rotation dimensions.} 
Results include the number of trained parameters and the corresponding errors. ``Avg.'' denotes the average Mean Absolute Error computed across ten standard planes. 
\dag indicates that these methods use the pre-trained encoder from DeiT.
}
\label{tab1:combined_results}
\begin{center}
\resizebox{0.8\textwidth}{!}{
\begin{tabular}{p{3.6cm}p{3.5cm}ccc}
\toprule
    \multirow{2}{*}{Setting} &
    \multirow{2}{*}{Method} &
    Trained &
    Translation &
    Rotation \\
    & & Params & Avg. (mm) & Avg. ($^\circ$) \\
\midrule
\midrule

    \multirow{2}{*}{\begin{tabular}[c]{@{}l@{}}Single-frame,\\ ImageNet-pretrained \end{tabular}} &
    DeiT \cite{touvron2021training} & 22.60M & 8.43 & 8.84 \\
    & DINOv2 \cite{oquab2023dinov2} & 90.28M & 8.22 & 8.72 \\
\midrule

    \multirow{7}{*}{\begin{tabular}[c]{@{}l@{}}Single-frame,\\ Pre-trained on \\ ultrasound data\end{tabular}} &
    BiomedCLIP \cite{zhang2023biomedclip} & 89.50M & 8.44 & 9.03 \\
    & LVM-Med \cite{mh2024lvm} & 89.44M & 8.56 & 8.90 \\
    & US-MoCo \cite{chen2021empirical} & 22.60M & 8.51 & 8.80 \\
    & US-IJEPA \cite{assran2023self} & 22.59M & 8.35 & 8.67 \\
    & US-MAE \cite{he2022masked} & 22.60M & 8.26 & 8.66 \\
    & USFM \cite{jiao2024usfm} & 92.94M & 8.26 & 8.62 \\
    & EchoCLIP \cite{christensen2024vision} & 89.74M & 8.21 & 8.52 \\
\midrule

    \multirow{2}{*}{\begin{tabular}[c]{@{}l@{}}Sequential,\\ ImageNet-pretrained \end{tabular}} &
    US-GuideNet$^{\dag}$ \cite{droste2020automatic} & 22.05M & 7.53 & 7.83 \\
    & Decision-T$^{\dag}$ \cite{chen2021decision} & 22.27M & 7.38 & 7.88 \\
\midrule

    \multirow{6}{*}{\begin{tabular}[c]{@{}l@{}}Sequential,\\ Pre-trained on \\ ultrasound data\end{tabular}} &
    BiomedCLIP & 86.15M & 7.29 & 7.80 \\
    & \cellcolor{gray!20}\textbf{BiomedCLIP+Ours} 
    & \cellcolor{gray!20}\textbf{3.94M} 
    & \cellcolor{gray!20}\textbf{5.55} 
    & \cellcolor{gray!20}\textbf{6.69} \\
    
    & USFM & 87.04M & 7.15 & 7.81 \\
    
    & \cellcolor{gray!20}\textbf{USFM+Ours} 
    & \cellcolor{gray!20}\textbf{3.97M} 
    & \cellcolor{gray!20}\textbf{5.35} 
    & \cellcolor{gray!20}\textbf{6.71} \\
    
    & EchoCLIP & 88.41M & 6.56 & 7.66 \\
    
    & \cellcolor{gray!20}\textbf{EchoCLIP+Ours} 
    & \cellcolor{gray!20}\textbf{2.61M} 
    & \cellcolor{gray!20}\textbf{5.40} 
    & \cellcolor{gray!20}\textbf{6.74} \\
\bottomrule
\end{tabular}
}
\end{center}
\end{table*}

\begin{figure*}[t]
\centering
\includegraphics[width=\textwidth]{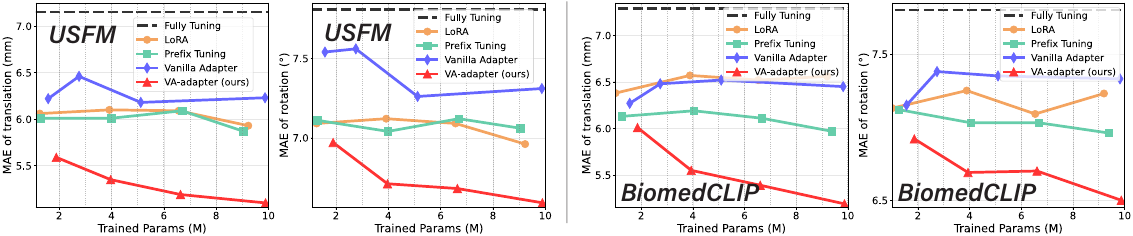}
    \caption{Performance comparison of PEFT methods on USFM and BiomedCLIP.}
    \label{peft}
\end{figure*}

\begin{figure*}[t]
\centering
\includegraphics[width=\textwidth]{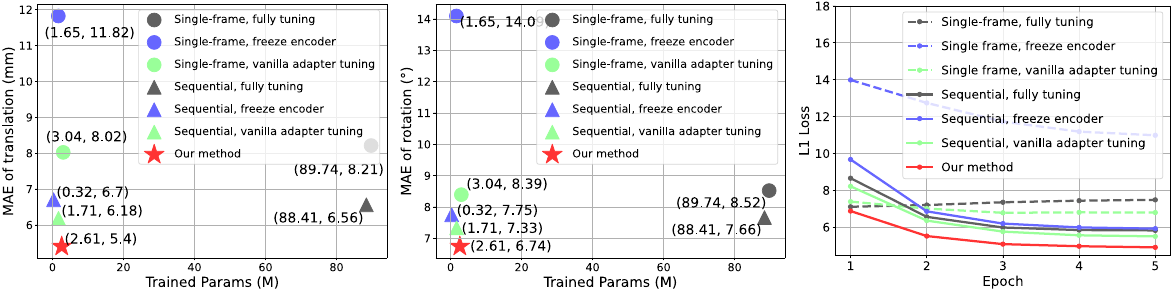}
    \caption{
    Ablation on vision-action interaction module of the EchoCLIP model.
    }
    \label{fig2:diff_train_method}
\end{figure*}

\subsection{Comparison with Baselines}
We evaluated our method on ten standard views, as shown in \Cref{tab1:combined_results}. 
Our method outperforms baselines in both MAE and parameter efficiency. 
Single-frame models rely only on the current image and cannot capture structural variation, leading to poor performance. 
Baseline sequential models~\cite{chen2021decision,droste2020automatic} typically fuse image and action features only in the prediction head, underusing structural cues in the sequence, and require full fine-tuning with high training cost. 
In contrast, our lightweight VA-Adapter is inserted into the image encoder, enabling progressive structure learning during feature extraction while improving both efficiency and accuracy.

We also compare with other common PEFT methods, including LoRA~\cite{DBLP:journals/corr/abs-2106-09685} and Prefix Tuning~\cite{li2021prefixtuningoptimizingcontinuousprompts}. Since they are designed for transformer-based backbones, we evaluate them only on USFM and BiomedCLIP. As shown in \cref{peft}, VA-Adapter performs best. This is likely because VA-Adapter explicitly models vision--action interactions to capture cardiac structure, whereas other PEFT methods mainly improve efficiency without enhancing structural understanding.

\begin{figure*}[t]
\centering
\includegraphics[width=\textwidth]{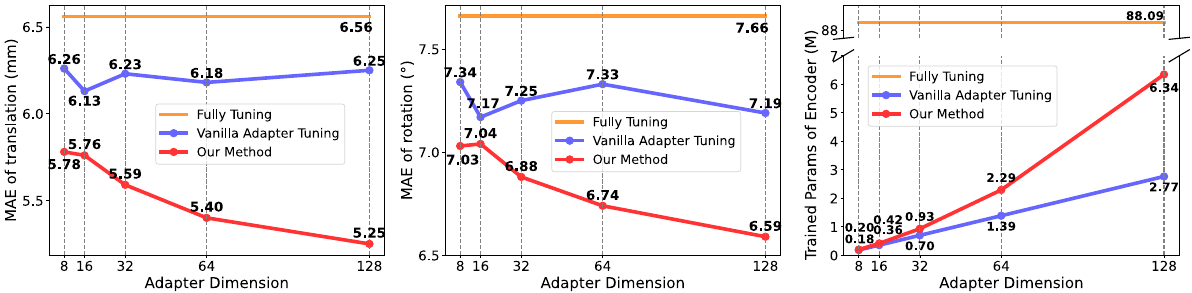}
    \caption{Ablation study on adapter dimension of the EchoCLIP model.}
    \label{fig3:adapter_dim}
\end{figure*}

\begin{figure*}[!t]
\includegraphics[width=\textwidth]{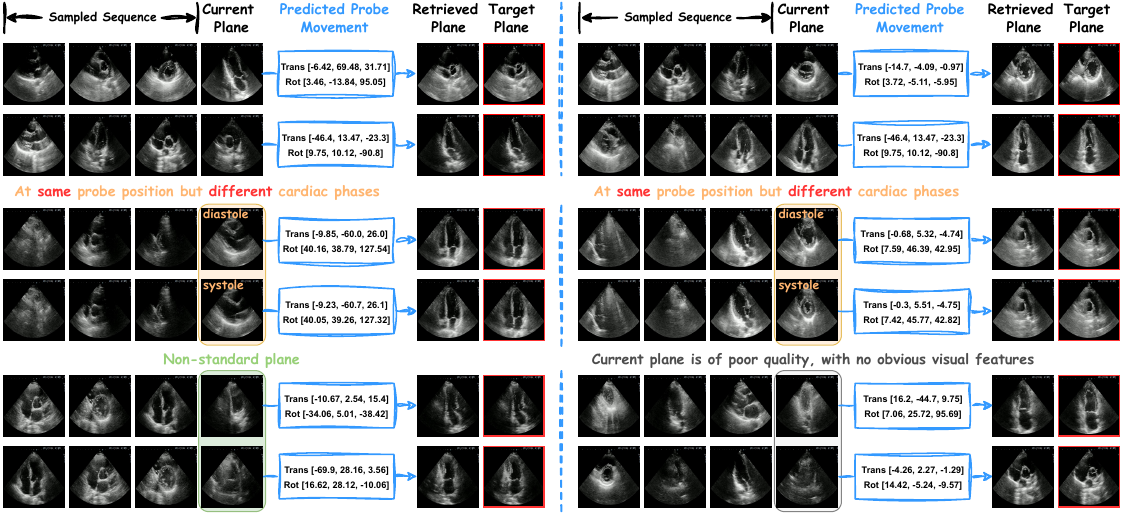}
\caption{Visualization of the model’s prediction.} \label{Vis}
\end{figure*}

\subsection{Ablation Study}

\textbf{Vision-action Interaction Module.}
We compare seven training baselines for EchoCLIP in \cref{fig2:diff_train_method}. The vanilla adapter removes the vision--action interaction module while keeping other settings unchanged. With 2.61M trainable parameters, our method achieves the lowest MAE and the best parameter--accuracy trade-off. Compared with the vanilla adapter of similar scale, it reduces MAE by 12.6\% / 8.0\% (Trans./Rot.) with only 0.9M extra parameters, demonstrating that the performance gains from the vision-action interaction mechanism outweigh the marginal cost of extra parameters. Furthermore, it also demonstrates faster convergence, outperforming other methods after the first epoch.

\textbf{Adapter Dimension.}
We study the effect of adapter dimension on performance and trainable parameters (\cref{fig3:adapter_dim}). With dimension $r{=}8$, our method introduces only 0.2M parameters (0.23\% of full fine-tuning) yet reduces MAE by 11.9\% / 8.2\% (Trans./Rot.), whereas the vanilla Adapter with similar size achieves only 4.6\% / 4.2\%. This indicates that our VA-Adapter captures cardiac structural cues more effectively under tight parameter budgets.
As $r$ increases from 8 to 128, our parameters grow by 31.7$\times$ while MAE further drops by 9.2\% / 6.3\%. In contrast, the vanilla adapter grows by 15.4$\times$ with only 0.2\% / 2.0\% additional MAE reduction, suggesting that our interaction module utilizes added parameters more efficiently.

\subsection{Visualization}

We visualize the model outputs in \cref{Vis}. We apply the predicted action to obtain the resulting pose, then perform nearest-neighbor retrieval in the scan sequence and compare the retrieved plane with the target. Results show that the model guides the probe toward the target plane. The outputs are consistent across frames from the same probe position but different cardiac phases, indicating robustness to the cardiac cycle. Moreover, for non-standard or low-quality planes with weak visual cues, the sequence model can still infer correct actions from vision–action relationships, which single-frame models cannot.

\subsection{Inference Real-time Analysis}
We benchmark inference on RTX 3090. Across all backbones, a single sequence takes $\sim$10.0--11.1\,ms without VA-Adapter and $\sim$10.5--11.8\,ms with VA-Adapter, satisfying real-time requirements. The added latency is marginal, indicating VA-Adapter preserves deployment efficiency in time-sensitive ultrasound guidance.

\section{Conclusion}

In this paper, we propose VA-Adapter, a lightweight module that empowers ultrasound  foundation models with probe guidance capability by modeling vision-action interactions during feature encoding. 
By combining basic knowledge from the foundation model with personalized 3D cardiac structures learned from the VA-Adapter, our method updates 95.4\%–97.0\% fewer parameters and reduces guidance error by 12.0\%–25.2\%, with extensive experiments validating superior efficiency and performance over state-of-the-art baselines.
This superior performance supports practical clinical scenarios, such as offering real-time assistance to junior sonographers and serving as the decision-making core for fully autonomous robotic ultrasound systems.

\bibliographystyle{splncs04}
\bibliography{reference}

\begin{thebibliography}{10}
\providecommand{\url}[1]{\texttt{#1}}
\providecommand{\urlprefix}{URL }
\providecommand{\doi}[1]{https://doi.org/#1}

\bibitem{assran2023self}
Assran, M., Duval, Q., Misra, I., Bojanowski, P., Vincent, P., Rabbat, M., LeCun, Y., Ballas, N.: Self-supervised learning from images with a joint-embedding predictive architecture. In: Proceedings of the IEEE/CVF Conference on Computer Vision and Pattern Recognition. pp. 15619--15629 (2023)

\bibitem{bao2024real}
Bao, M., Wang, Y., Wei, X., Jia, B., Fan, X., Lu, D., Gu, Y., Cheng, J., Zhang, Y., Wang, C., et~al.: Real-world visual navigation for cardiac ultrasound view planning. In: International Conference on Medical Image Computing and Computer-Assisted Intervention. pp. 317--326. Springer (2024)

\bibitem{chen2021decision}
Chen, L., Lu, K., Rajeswaran, A., Lee, K., Grover, A., Laskin, M., Abbeel, P., Srinivas, A., Mordatch, I.: Decision transformer: Reinforcement learning via sequence modeling. Advances in neural information processing systems  \textbf{34},  15084--15097 (2021)

\bibitem{chen2021empirical}
Chen, X., Xie, S., He, K.: An empirical study of training self-supervised vision transformers. In: Proceedings of the IEEE/CVF international conference on computer vision. pp. 9640--9649 (2021)

\bibitem{christensen2024vision}
Christensen, M., Vukadinovic, M., Yuan, N., Ouyang, D.: Vision--language foundation model for echocardiogram interpretation. Nature Medicine pp.~1--8 (2024)

\bibitem{droste2020automatic}
Droste, R., Drukker, L., Papageorghiou, A.T., Noble, J.A.: Automatic probe movement guidance for freehand obstetric ultrasound. In: Medical Image Computing and Computer Assisted Intervention--MICCAI 2020: 23rd International Conference, Lima, Peru, October 4--8, 2020, Proceedings, Part III 23. pp. 583--592. Springer (2020)

\bibitem{ghorbani2020deep}
Ghorbani, A., Ouyang, D., Abid, A., He, B., Chen, J.H., Harrington, R.A., Liang, D.H., Ashley, E.A., Zou, J.Y.: Deep learning interpretation of echocardiograms. NPJ digital medicine  \textbf{3}(1), ~10 (2020)

\bibitem{he2022masked}
He, K., Chen, X., Xie, S., Li, Y., Doll{\'a}r, P., Girshick, R.: Masked autoencoders are scalable vision learners. In: Proceedings of the IEEE/CVF conference on computer vision and pattern recognition. pp. 16000--16009 (2022)

\bibitem{DBLP:journals/corr/abs-2106-09685}
Hu, E.J., Shen, Y., Wallis, P., Allen{-}Zhu, Z., Li, Y., Wang, S., Chen, W.: Lora: Low-rank adaptation of large language models. CoRR  \textbf{abs/2106.09685} (2021)

\bibitem{jiang2024structure}
Jiang, H., Li, M., Sun, Z., Jia, N., Sun, Y., Luo, S., Song, S., Huang, G.: Structure-aware world model for probe guidance via large-scale self-supervised pre-train. arXiv preprint arXiv:2406.19756  (2024)

\bibitem{jiang2024cardiac}
Jiang, H., Sun, Z., Jia, N., Li, M., Sun, Y., Luo, S., Song, S., Huang, G.: Cardiac copilot: Automatic probe guidance for echocardiography with world model. arXiv preprint arXiv:2406.13165  (2024)

\bibitem{jiang2025ultrasep}
Jiang, H., Wang, T., Sun, Z., Wang, Y., Yue, Y., Sun, Y., Jia, N., Li, M., Luo, S., Song, S., et~al.: Ultrasep: Sequence-aware pre-training for echocardiography probe movement guidance. Pattern Recognition p. 112600 (2025)

\bibitem{jiang2025towards}
Jiang, H., Zhao, A., Yang, Q., Yan, X., Wang, T., Wang, Y., Jia, N., Wang, J., Wu, G., Yue, Y., et~al.: Towards expert-level autonomous carotid ultrasonography with large-scale learning-based robotic system. Nature Communications  \textbf{16}(1), ~7893 (2025)

\bibitem{jiao2024usfm}
Jiao, J., Zhou, J., Li, X., Xia, M., Huang, Y., Huang, L., Wang, N., Zhang, X., Zhou, S., Wang, Y., et~al.: Usfm: A universal ultrasound foundation model generalized to tasks and organs towards label efficient image analysis. Medical Image Analysis  \textbf{96},  103202 (2024)

\bibitem{li2023rl}
Li, K., Li, A., Xu, Y., Xiong, H., Meng, M.Q.H.: Rl-tee: Autonomous probe guidance for transesophageal echocardiography based on attention-augmented deep reinforcement learning. IEEE Transactions on Automation Science and Engineering  \textbf{21}(2),  1526--1538 (2023)

\bibitem{li2021prefixtuningoptimizingcontinuousprompts}
Li, X.L., Liang, P.: Prefix-tuning: Optimizing continuous prompts for generation. In: Proceedings of the 59th Annual Meeting of the Association for Computational Linguistics and the 11th International Joint Conference on Natural Language Processing (Volume 1: Long Papers) (2021)

\bibitem{mh2024lvm}
MH~Nguyen, D., Nguyen, H., Diep, N., Pham, T.N., Cao, T., Nguyen, B., Swoboda, P., Ho, N., Albarqouni, S., Xie, P., et~al.: Lvm-med: Learning large-scale self-supervised vision models for medical imaging via second-order graph matching. Advances in Neural Information Processing Systems  \textbf{36} (2024)

\bibitem{mitchell2019guidelines}
Mitchell, C., Rahko, P.S., Blauwet, L.A., Canaday, B., Finstuen, J.A., Foster, M.C., Horton, K., Ogunyankin, K.O., Palma, R.A., Velazquez, E.J.: Guidelines for performing a comprehensive transthoracic echocardiographic examination in adults: recommendations from the american society of echocardiography. Journal of the American Society of Echocardiography  \textbf{32}(1),  1--64 (2019)

\bibitem{narang2021utility}
Narang, A., Bae, R., Hong, H., Thomas, Y., Surette, S., Cadieu, C., Chaudhry, A., Martin, R.P., McCarthy, P.M., Rubenson, D.S., et~al.: Utility of a deep-learning algorithm to guide novices to acquire echocardiograms for limited diagnostic use. JAMA cardiology  \textbf{6}(6),  624--632 (2021)

\bibitem{oquab2023dinov2}
Oquab, M., Darcet, T., Moutakanni, T., Vo, H., Szafraniec, M., Khalidov, V., Fernandez, P., Haziza, D., Massa, F., El-Nouby, A., et~al.: Dinov2: Learning robust visual features without supervision. arXiv preprint arXiv:2304.07193  (2023)

\bibitem{ouyang2020video}
Ouyang, D., He, B., Ghorbani, A., Yuan, N., Ebinger, J., Langlotz, C.P., Heidenreich, P.A., Harrington, R.A., Liang, D.H., Ashley, E.A., et~al.: Video-based ai for beat-to-beat assessment of cardiac function. Nature  \textbf{580}(7802),  252--256 (2020)

\bibitem{roth2017global}
Roth, G.A., Johnson, C., Abajobir, A., Abd-Allah, F., Abera, S.F., Abyu, G., Ahmed, M., Aksut, B., Alam, T., Alam, K., et~al.: Global, regional, and national burden of cardiovascular diseases for 10 causes, 1990 to 2015. Journal of the American college of cardiology  \textbf{70}(1),  1--25 (2017)

\bibitem{touvron2021training}
Touvron, H., Cord, M., Douze, M., Massa, F., Sablayrolles, A., J{\'e}gou, H.: Training data-efficient image transformers \& distillation through attention. In: International conference on machine learning. pp. 10347--10357. PMLR (2021)

\bibitem{wang2016temporal}
Wang, L., Xiong, Y., Wang, Z., Qiao, Y., Lin, D., Tang, X., Van~Gool, L.: Temporal segment networks: Towards good practices for deep action recognition. In: European conference on computer vision. pp. 20--36. Springer (2016)

\bibitem{wang2025ultrahit}
Wang, T., Jiang, H., Wang, Y., Sun, Z., Yan, X., Li, X., Huang, G.: Ultrahit: A hierarchical transformer architecture for generalizable internal carotid artery robotic ultrasonography. arXiv preprint arXiv:2509.13832  (2025)

\bibitem{yang2024mma}
Yang, L., Zhang, R.Y., Wang, Y., Xie, X.: Mma: Multi-modal adapter for vision-language models. In: Proceedings of the IEEE/CVF Conference on Computer Vision and Pattern Recognition. pp. 23826--23837 (2024)

\bibitem{yue2025echoworld}
Yue, Y., Wang, Y., Jiang, H., Liu, P., Song, S., Huang, G.: Echoworld: Learning motion-aware world models for echocardiography probe guidance. In: Proceedings of the Computer Vision and Pattern Recognition Conference. pp. 25993--26003 (2025)

\bibitem{zhang2023biomedclip}
Zhang, S., Xu, Y., Usuyama, N., Xu, H., Bagga, J., Tinn, R., Preston, S., Rao, R., Wei, M., Valluri, N., et~al.: Biomedclip: a multimodal biomedical foundation model pretrained from fifteen million scientific image-text pairs. arXiv preprint arXiv:2303.00915  (2023)

\end{thebibliography}

\end{document}